\documentclass{article}

\usepackage{PRIMEarxiv}

\usepackage[utf8]{inputenc} 
\usepackage[T1]{fontenc}    
\usepackage{hyperref}       
\usepackage{url}            
\usepackage{booktabs}       
\usepackage{amsfonts}       
\usepackage{nicefrac}       
\usepackage{microtype}      
\usepackage{lipsum}
\usepackage{fancyhdr}       
\usepackage{graphicx}       
\graphicspath{{media/}}     

\usepackage{amssymb}
\usepackage{algorithm}
\usepackage{algpseudocode}
\usepackage{amsmath}
\usepackage{graphicx}
\usepackage{wrapfig}
\usepackage{subcaption}
\usepackage{cite}
\usepackage{enumerate}
\usepackage{mathrsfs}

\title{TexIm FAST: Text-to-Image Representation for Semantic Similarity Evaluation using Transformers
}

\author{
  Wazib~Ansar \\
  A. K. Choudhury School of IT \\
  University of Calcutta \\
  Kolkata, West Bengal, India \\
  \texttt{waakcs\_rs@caluniv.ac.in} \\
   \And
  Saptarsi~Goswami \\
  Department of Computer Science \\
  Bangabasi Morning College \\
  Kolkata, West Bengal, India \\
  \texttt{sgakc@caluniv.ac.in} \\
    \And
  Amlan~Chakrabarti \\
  A. K. Choudhury School of IT \\
  University of Calcutta \\
  Kolkata, West Bengal, India \\
  \texttt{acakcs@caluniv.ac.in} \\  
}

\begin{document}
\maketitle

\begin{abstract}
One of the principal objectives of Natural Language Processing (NLP) is to generate meaningful representations from text. Improving the informativeness of the representations has led to a tremendous rise in the dimensionality and the memory footprint. It leads to a cascading effect amplifying the complexity of the downstream model by increasing its parameters. The available techniques cannot be applied to cross-modal applications such as text-to-image. To ameliorate these issues, a novel Text-to-Image methodology for generating fixed-length representations through a self-supervised Variational Auto-Encoder (VAE) for semantic evaluation applying transformers (TexIm FAST) has been proposed in this paper. The pictorial representations allow oblivious inference while retaining the linguistic intricacies, and are potent in cross-modal applications. TexIm FAST deals with variable-length sequences and generates fixed-length representations with over 75\% reduced memory footprint. It enhances the efficiency of the models for downstream tasks by reducing its parameters. The efficacy of TexIm FAST has been extensively analyzed for the task of Semantic Textual Similarity (STS) upon the MSRPC, CNN/ Daily Mail, and XSum data-sets. The results demonstrate 6\% improvement in accuracy compared to the baseline and showcase its exceptional ability to compare disparate length sequences such as a text with its summary.
\end{abstract}

\keywords{Oblivious Inference \and Sequence Embedding \and Text-to-Image \and Transformers NLP \and Variational Auto-Encoder}

\section{Introduction}
Text is a ubiquitous mode of communication, knowledge dissemination, and information preservation. It is the most convenient media to navigate and retrieve information concerning a topic from voluminous repositories \cite{Ref1}. The steady progress in the field of Natural Language Processing (NLP) has enabled seamless comprehension and generation of unstructured texts with high precision \cite{Ref2}. Considering the colossal amount of text at our disposal, research directed towards devising techniques for effective representation of unstructured textual content has become all the more essential \cite{Ref3}. Despite abundant research, most of them rely on generating multi-dimensional vector representations of text leading to the repercussions associated with the "curse of dimensionality" like huge memory footprint and exorbitant computational complexity of downstream tasks \cite{Ref4}.

The text representation techniques are often domain-specific, i.e. the vector representation of a text sequence cannot be fed into an image-specific model. To bridge this gap between text and image, research must be directed towards formulating cross-modal approaches to generate image representations from text comprising the linguistic intricacies. A few studies have been conducted to replace text with equivalent images \cite{Ref5}, or to produce modified illustrations \cite{Ref6, Ref7, Ref8}. But, none of these generate pictorial representations from textual embeddings that can equivalently be applied to both image and text-oriented models for downstream tasks.

Semantic Textual Similarity (STS) is one of the renowned tasks in NLP that revolves around assessing the level of semantic equivalence between pairs of sequences as a continuous value \cite{Ref47}. STS plays a pivotal role in NLP applications like Automatic Text Summarization (ATS) \cite{Ref42}, Plagiarism Detection (PD) \cite{Ref43}, Question Answering (QA) \cite{Ref41}, and Machine Translation (MT) \cite{Ref40}. STS goes beyond superficial similarity considering factors such as the overlap, order, and the length of common sub-sequences \cite{Ref44, Ref45, Ref46}. Rather it relies on ontology \cite{Ref49, Ref54}, context \cite{Ref55}, semantics \cite{Ref56, Ref57} accompanied with deep learning based architectures \cite{Ref58, Ref59, Ref60, Ref61}. Although such approaches provide satisfactory results, they succumb while dealing with disparate sequence lengths such as comparing a text with its associated summary.

To address the research gaps stated above, we direct our research towards devising an innovative "Text-to-Image" (TexIm) encoding. These cross-modal encodings are memory-efficient, can be visualized as an image, and assimilate the linguistic intricacies of the input text. This paper is an improvement over the contribution by Ansar et al. \cite{Ref11} whereby they generated informed pictorial representations of texts capturing the contextualized syntactic and lexical features using BERT with 37.37\% reduced memory footprint. However, this approach had some drawbacks, namely- (1) the size of the representations varied according to sequence lengths as it assigned each word as a pixel requiring padding to ensure uniform lengths, (2) this reduced its effectiveness while comparing representations of disparate sequence lengths, (3) the Principal Component Analysis (PCA) applied for dimension reduction led to loss of information. To mitigate these shortcomings, we hereby put forth our endeavor coined as \textbf{TexIm FAST} for Text-to-Image conversion with \underline{F}ixed-length representations applying \underline{A}utoencoder for \underline{S}emantic evaluation through \underline{T}ransformers. It includes a novel Variational Auto-Encoder (VAE) based on Convolutional Neural Network (CNN) and Transformer with Selective Learn-Forget Network (TSLFN) to encode the input text into fixed-length pictorial representations occupying less than 75\% memory. Furthermore, a weighted annealing mechanism is applied to the VAE to ameliorate posterior collapse wherein the latent variable becomes uninformative and gets ignored by the decoder \cite{Ref72}. Compared to the previous TexIm rendition- (1) It encodes the entire sequence as a whole instead of word-level representations. (2) It eliminates the bias infused due to variable-length sequences. (3) It generates the pixels using a self-supervised VAE eliminating the need for any annotated data-sets. Besides, it allows oblivious inference through encoded pictorial representations that can be utilized for further analysis without revealing the actual text \cite{Ref69}. As a downstream task, a novel architecture utilizing TSLFN has been proposed for STS. Its performance on STS demonstrates its exceptional ability to deal with disparate-length sequences. The principal contributions of this paper are as follows:

\begin{enumerate}
    \item We propose TexIm FAST, an improved Text-to-Image encoding technique that allows oblivious inference retaining the semantic information in the text with over 75\% reduction in memory footprint.
    
    \item It comprises a self-supervised CNN-TSLFN based VAE to capture the contextualized semantic information in a sequence holistically and encode it into a uniform-dimensional image irrespective of its length.

    \item The efficacy of the pictorial representations can be observed through 5.6\% higher accuracy over the baseline for STS task applying a novel TSLFN architecture.

\end{enumerate}

 The remainder of this paper has been outlined herein. \textit{Section \ref{relw}} presents the literature surveyed. \textit{Section \ref{pm}} enunciates the methodology of the proposed TexIm FAST. \textit{Section \ref{exp}} provides the details of the experiment conducted. \textit{Section \ref{rdis}} showcases the results obtained through extensive evaluation. Finally, the conclusions are drawn and the future scope is discussed in \textit{Section \ref{cocl}}.

\section{Related Works}
\label{relw}
In this section, a review of the preceding works has been undertaken to assess the inherent gaps in research. For ease of comprehension, the commentary of works has been grouped under subsections dedicated to cross-modal representations and semantic similarity.

\subsection{Cross-Modal Representation}
\label{vist}

The related works can be segregated into cross-modal matching as-well-as generation of cross-modal representations. For cross-modal matching, Yang et al. \cite{Ref81} developed a quantization algorithm with shared predictive-representations and label alignment. Xu et al. \cite{Ref82} proposed cross-modal matching with co-attention-based alignment of image regions with words in textual descriptions along with global semantic prediction of labels. Jin et al. \cite{Ref83} devised a CNN-based cross-modal matching model with a specialized hashing technique exploiting spatial information to target efficiency. 

A few notable works on cross-modal generation have been discussed herein. Zakraoui et al. \cite{Ref5} substituted text with the most probable illustrations from a repository. Das et al. \cite{Ref30} applied stack Generative Adversarial Network (GAN) and Long Short-Term Memory (LSTM)-based autoencoders. Ramesh et al. \cite{Ref32} formulated a zero-shot learning-based transformer. Saharia et al. \cite{Ref31} utilized pre-trained language models for encoding text accompanied by a diffusion model for image generation. Gu et al. \cite{Ref33} constructed a VAE with a quantized vector in association with a diffusion model. Tan et al. \cite{Ref84} generated images from textual captions applying GAN with spatial attention and VAE for normalized latent distributions. In these works, the generated illustrations are devoid of the linguistic characteristics of the original text. This limits their application to downstream NLP tasks. 

Besides, a handful of research has been conducted to generate visual representations of text capturing its linguistic intricacies. Nataraj et al. \cite{Ref6} converted text into grayscale images having pixels with varying intensity values corresponding to the characters. The images were reported to be immune to API injection and obfuscation of code for the application of malware detection. He et al. \cite{Ref7} further improved it through Spatial Pyramid Pooling (SPP) to accommodate inputs having variable lengths. Petrie et al. \cite{Ref8} focused on string-map visualizations for patent-inventor record comparison. From the above-mentioned contributions, it can be inferred that textual content can be transformed into cross-modal representations flexible enough to be processed either as text or as images.

\subsection{Semantic Similarity}
\label{ssim}
The primeval works on evaluating the similarity between texts focused on identifying common sub-sequences. A few such similarity measures are even prevalent in the efficacy evaluation of summarization or translation models in NLP. Chin-Yew Lin \cite{Ref46} proposed a recall-oriented metric relying on the overlap of n-grams to determine the similarity among a pair of texts. Contrastingly, Papineni et al. \cite{Ref45} computed the similarity through a precision-oriented metric with a penalty term to account for disparate sequence lengths. However, these approaches relied on superficial factors including overlap, order, and common sub-sequence length \cite{Ref44}.

Besides identifying common sub-sequences, STS takes into account ontology, context as-well-as semantics of sequences. Sánchez and Batet \cite{Ref49} calculated through the context of terms inferred from the ontological knowledge base. To mitigate the restriction imposed by predefined ontologies, Jiang et al. \cite{Ref54} formulated a feature-oriented approach utilizing Wikipedia as the knowledge base. The Cilibrasi and Vitanyi \cite{Ref55} calculated the similarity of words by constructing the context depending on Google search results. Kusner et al. \cite{Ref56} calculated semantic similarity as the minimum distance between the embeddings of a word in one sequence and its nearest word in the other sequence. Wang et al. \cite{Ref57} decomposed the sentences for comparison into words and calculated the cosine similarity among them. The recent trend reflects the inclination towards deep learning models. He and Lin \cite{Ref58} combined Bi-LSTM with CNN for modeling the context followed by a pair-wise comparison of the hidden states. Tien et al. \cite{Ref59} proposed an LSTM-based similarity prediction architecture utilizing sentence embeddings constructed by a CNN-LSTM architecture. Lopez-Gazpio et al. \cite{Ref60} formulated an attention-based model taking n-grams as inputs to construct attention vectors for the sentence pairs and fed them to an aggregation layer. Li et al. \cite{Ref61} proposed a cross-self attention-based Siamese Neural Network for determining the semantic similarity of biomedical texts. A common drawback of STS approaches is that they falter with disparate-length sequences.

These developments show that cross-modal representations expand the horizons for processing data. Besides, techniques like quantization, hashing, or latent representation generation facilitate memory-efficiency. Moreover, semantic similarity is one of the prominent tasks in NLP to assess the efficacy of textual representations. However, devising a light-weight model capable enough to capture the linguistic intricacies of cross-modal representations with disparate length sequences effectively is a challenging task.

\section{Proposed Methodology}
\label{pm}
The proposed methodology has been segregated into two coherent sub-sections- \textit{Section \ref{tfas}} describes the proposed TexIm FAST for the generation of pictorial representations from the text. While \textit{Section \ref{stsa}} presents the novel TSLFN-based architecture for evaluation of the TexIm FAST representations upon the downstream task of STS.
    
\subsection{TexIm FAST}
\label{tfas}
 The flow steps in the proposed TexIm FAST have been elucidated herein-below:
 
\subsubsection{Pre-processing}
\label{prep}
 The input text is first put through cleaning processes to remove irrelevant portions like line breaks, tabulations, URL patterns, numbers, non-ASCII characters and punctuations. Standardization of white space is performed.
 \bigskip

   \begin{algorithm}
    \caption{Tokenization Lookup Table Generation }
    \label{alg:rfmg}
    \begin{algorithmic}
    \Require $\textrm{Vocabulary } V$ 
    \Ensure $\textrm{ Tokenization Lookup } T_{id}$

    \State $l_{V} \gets |V|$ \Comment{$l_{V}$ is the size of $V$}
    \For{$i \gets 0$ to $(l_{V}-1)$}      
        \State $T_{id}[i] \gets \{V_{i}:t[i]\}$
        \Comment{$V_{i}$ is mapped to $t[i]$}
    \EndFor

    \end{algorithmic}
    \end{algorithm}

    \smallskip

\subsubsection{Tokenization}
\label{tzn}
The pre-processed text undergoes sub-word tokenization wherein a word is split into meaning preserving morphemes \cite{Ref38}. This reduces the vocabulary size by allowing different words to be formed through the combination of a given set of sub-words. For instance, words like "walk", "walking", "talk" and "talking" can be tokenized into ["walk", "talk" and "\#\#ing"]. Moreover, it improves comprehension by rendering meaning to its constituent sub-words. For instance, from the tokens like "snow", "man", and "fire", words like "snow", "snowman", "fire" and "fireman" can be formed. A lookup table is constructed wherein each sub-word in the vocabulary $V$ is assigned a unique integer identifier $t[i]$ ranging from $1$ to $|V|$ as portrayed in Algorithm \ref{alg:rfmg}. Relying on it, a tokenized sequence of integers $S_{t}$ is obtained such that $V_{i}$ in the sequence is replaced with its corresponding integer identifier $t_{id}[i]$. To ensure uniform sequence length $L$, padding token(s) are appended as depicted in equation (1).

    \begin{flushleft}
     \[ S_{t}[i] = \begin{cases}
          t_{id}[i] & \text{for sub-word tokens}\\
          0 & \text{for padding}
        \end{cases}    
        \textrm{, } \forall{i \in {[0, sl)}} \hspace*{30pt}\hfill (1)
    \] 
    \end{flushleft}

    \smallskip

\subsubsection{Input Embedding}
\label{irep}

The text is then transformed to a sequence of embeddings $S_{e}$ such that each token $S_{t}[i]$ in the sequence is replaced with a unique vector $e[i]$. The embeddings are randomly initialized as $d$ dimensional vectors and optimized during training just as other network parameters. For contextualized representation, position embedding is applied to the sequence to establish the ordering of the tokens. For this, the positional index $S_{p}[i]$ of the $i^{th}$ token in the sequence is determined using equation (2). 

    \begin{flushleft}
     \[ S_{p}[i] = \begin{cases}
          \theta + i\delta & \text{for sub-word tokens}\\
          -1 & \text{for padding}
        \end{cases}    
        \textrm{, } \forall{i \in {[0, sl)}} \hspace*{27pt}\hfill (2)
    \] 
    \end{flushleft}

    \smallskip

where $\delta$ is the increment value with $\theta$ serving as the offset. The $d$ dimensional position embeddings are then learnt during network training utilizing $S_{p}$. Finally, the tokenized integer sequence $S_{t}$ and the position encoded sequence $S_{p}$ are element-wise concatenated (denoted by $\oplus$) to obtain the embedded sequence $S \in \mathbb{R}^{L \times D}$ as shown in equation (3).
    \newline 
    
    \begin{flushleft}
    $S= S_{p} \oplus S_{t}$ \hspace*{0pt}\hfill (3)
    \end{flushleft}

\smallskip

 \begin{figure}[h]
\centering
\includegraphics[width=0.6\linewidth, height=14.4cm]{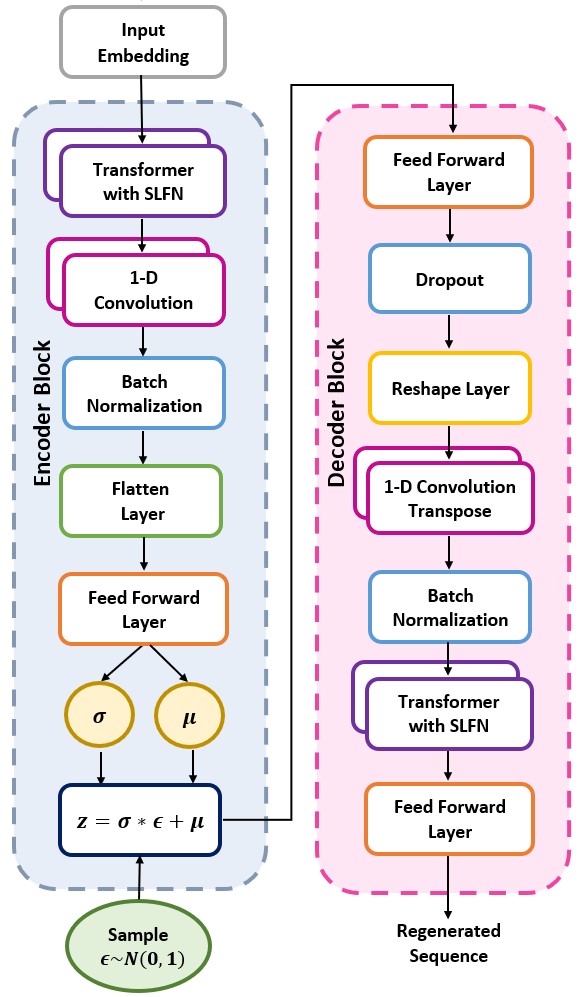}
\caption{Illustration of the proposed TexIm FAST VAE architecture for projection of sequences to fixed-length vector}
\label{fig2}       
\end{figure}

\subsubsection{Low-Dimensional Projection}
\label{svpr}
The embedded input sequence is transformed to a low-dimensional fixed-length vector from which the TexIm FAST encodings can be derived. To accomplish this, a novel VAE architecture based on CNN and TSLFN has been formulated as depicted in Figure \ref{fig2}. The model is trained through a self-supervised learning mechanism wherein it takes reference for the output from the input itself \cite{Ref66}. A merit of the proposed VAE is that it encodes any variable-length sequence into a fixed-length representation. This helps to generate images of equal resolution irrespective of their input sequence lengths. The salient features of the model have been presented as follows:
\begin{enumerate}[(i)]
    \item \textbf{Transformer with SLFN (TSLFN)}: It comprises of Multi-Head Attention (MHA) followed by a Selective Learn-Forget Network (SLFN) as depicted in Figure \ref{fig12}. The MHA $Att^{m}$ concatenates $H$ self-attention units $Att^{s}$ projected through $W^{o} \in \mathbb{R}^{D} \times \mathbb{R}^{D}$ to accommodate parallel computation involving diverse positions \cite{Ref67} as expressed in equations (4) and (5).
    \newline
    
    \begin{flushleft}
    $Att^{s}=softmax(\frac{Q_{s}K_{s}^{T}}{\sqrt{D}})V_{s}$ \hspace*{0pt}\hfill (4) \linebreak
    \end{flushleft}
    
    \begin{flushleft}
    $Att^{m}=Concatenate(Att^{s}_{i})W^{o}, \forall{i \in H}$ \hspace*{0pt}\hfill (5) \linebreak
    \end{flushleft}

    where, query $Q_{s}$, key $K_{s}$ and value $V_{s}$ are calculated through linear transformations on the sequence $Z \in \mathbb{R}^{L} \times \mathbb{R}^{D}$ along with weight matrices $W^{q}, W^{k}$ and $W^{v}$ $ \in \mathbb{R}^{D \times \frac{D}{H}}$ as formulated in equation (6).
    \newline
    
    \begin{flushleft}
    $Q_{s},K_{s},V_{s}=W^{q}Z,W^{k}Z,W^{v}Z$ \hspace*{0pt}\hfill (6) \linebreak
    \end{flushleft}

    The SLFN \cite{Ref62} augments the efficacy of the residual connections by effectively capturing the dependencies. It filters insignificant information reducing the overall complexity. It achieves selective learning through a gated mechanism involving a combination of sigmoid $Sg_{i}$ and tanh $Tg_{i}$ gates to update the hidden state $H_{i}$ as shown in equation (7).
    \newline
    
    \begin{flushleft}
    $H_{i}=Sg_{i}+(Sg_{i} \odot Tg_{i})$ \hspace*{0pt}\hfill (7) \linebreak
    \end{flushleft}

    given that, 
    
    \begin{flushleft}
    $Sg_{i}=\sigma[(X_{i}W^{s}_{i})+(H_{i-1}U^{s}_{i})]$ \hspace*{0pt}\hfill (8) \linebreak
    \end{flushleft}

    \begin{flushleft}
    $Tg_{i}=tanh[(X_{i}W^{t}_{i})+(H_{i-1}W^{t}_{i})]$ \hspace*{0pt}\hfill (9) \linebreak
    \end{flushleft}

    \begin{figure}[h]
    \centering
    \includegraphics[width=0.39\linewidth]{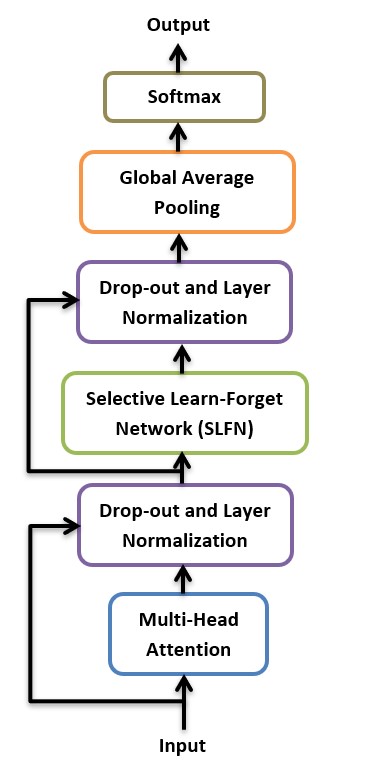}
    \caption{Illustration of the TSLFN architecture \cite{Ref62}}
    \label{fig12}       
    \end{figure}

    where, $X_{i}$, $H_{i-1}$, $W^{s}_{i}$ and $U^{s}_{i}$ denote the $i^{th}$ step input, previous step hidden state, input weight matrix and the hidden state weight matrix respectively.

    \item \textbf{1-D Convolution:} It applies multiple linear filters over a sequence in a feed-forward fashion to extract meaningful features through sparse interactions and parameter sharing \cite{Ref70}. Given a sequence $X \in \mathbb{R}^{L \times D}$, a filter $F_{c} \in \mathbb{R}^{ F \times D} $ performs successive element-wise products over subsets $S_{f} \in \mathbb{R}^{ F \times D}$; $ \forall S_{f} \in X $ to generate a feature map $ F_{m} \in \mathbb{R}^{ (L-F+1) \times D}$ as portrayed in equation (10).
    \newline
    
    \begin{flushleft}
    $F_{m}[i]= F_{c} \odot X_{i:i+F-1} \textrm{, } \forall{i \in [0, (L-F)]} $ \hspace*{0pt}\hfill (10) \linebreak
    \end{flushleft}
    
    where $X_{i:j}$ denotes a sub-set of $X$ from position $i$ to $j$.

    \item \textbf{Reparameterization and Objective Function:} Utilizing the mean $\mu$ and standard deviation $\sigma$ produced by the encoder, a Gaussian distribution is generated. A latent vector $z$ is sampled from it as shown in equation (11).
    \newline
    
    \begin{flushleft}
    $ z \sim \mathscr{N}(\mu, \sigma)$ \hspace*{0pt}\hfill (11) \linebreak
    \end{flushleft}

    To make the sampling process continuous and differentiable, reparameterization is performed sampling $\epsilon$ from a standard Gaussian distribution, i.e. $\epsilon \sim \mathscr{N}(0, 1)$ and recomputing $z$ as follows \cite{Ref71}:
    \newline
    
    \begin{flushleft}
    $ z = \sigma * \epsilon + \mu$ \hspace*{0pt}\hfill (12) \linebreak
    \end{flushleft}

    For training the network, Evidence Lower Bound (ELBo) $J_{\text{VAE}}(\theta, \phi; x)$ \cite{Ref71} has been considered the objective function. It consists of the reconstruction loss $\mathbb{E}_{z \sim q(z | x; \phi)}\left[\log p(x | z; \theta)\right]$ and the Kullback-Leibler (KL) divergence $\text{D}_{\text{KL}}()$ between the encoded distribution $q(z | x; \phi)$ and the required prior distribution $p(z)$ as shown in equation (13) with $x$, $\phi$ and $\theta$ denoting the input, the encoder and the decoder parameters respectively.
     \newline
    
    \begin{flushleft}
    $J_{\text{VAE}}(\theta, \phi; x) = \text{D}_{\text{KL}}\left(q(z | x; \phi) \,||\, p(z)\right)$ \\
    $ - \mathbb{E}_{z \sim q(z | x; \phi)}\left[\log p(x | z; \theta)\right]$ \hspace*{0pt}\hfill (13) \linebreak
    \end{flushleft}

    To avoid posterior collapse, a varying weight term $W_{a}$ increasing with the training epochs is multiplied to $\text{D}_{\text{KL}}\left(q(z | x; \phi) \,||\, p(z)\right)$ as follows:
    \newline
    
    \begin{flushleft}
    $W_{a}=(1+e^{-(\frac{n}{N}+b)})^{-1}$ \hspace*{0pt}\hfill (14) \linebreak
    \end{flushleft}

    where, $n, N, b \in \mathbb{N}$ denote the current epoch, total number of epochs and the control parameter. This is analogous to the annealing process and lays more emphasis on reconstruction loss. The modified ELBo $J'_{\text{VAE}}(\theta, \phi; x)$ has been portrayed in equation (15).
    \newline
    
    \begin{flushleft}
    $J'_{\text{VAE}}(\theta, \phi; x) = W_{a}*\text{D}_{\text{KL}}\left(q(z | x; \phi) \,||\, p(z)\right)$ \\
    $- \mathbb{E}_{z \sim q(z | x; \phi)}\left[\log p(x | z; \theta)\right]$ \hspace*{0pt}\hfill (15) \linebreak
    \end{flushleft}
\end{enumerate}

\smallskip

\subsubsection{Normalization and Feature Scaling}
\label{nfs}
The fixed-length sequence embedding $E$ having $dim_e$  dimensions obtained in the preceding step requires additional processing to be considered as pixel intensity values. The components are normalized within the range of [0, 1] to achieve this (equation (16)). These normalized embeddings $E'$ are then scaled between acceptable grayscale image intensity ranges, i.e. [0, 255] and $E^{s}$ is obtained as shown in equation (17).
    \newline
    \begin{flushleft}
    $E'= \frac{e_{i}-min(E)}{max(E)-min(E)}\textrm{, } \forall{e_{i} \in E}$ \hspace*{0pt}\hfill (16) 
    \end{flushleft}

    \smallskip

    \begin{flushleft}
    $E^{s}=e'_{i}*255\textrm{, } \forall{e'_{i} \in E'}$ \hspace*{0pt}\hfill (17) 
    \end{flushleft}

 \smallskip

     \begin{algorithm}
    \caption{Reshaping the Embedding Sequence}
    \label{alg:rsem}
    \begin{algorithmic}
    \Require $\textrm{Scaled Embedding Sequence } E^{s} \textrm{ of length }dim_e$
    \Ensure $I_{l} \times I_{b} \textrm{ dimensional Embedding Matrix } E^{G}$
    \State $E^{G} \gets I_{l} \times I_{b} \textrm{ dimensional matrix initialized with zeros}$
    \State $k \gets 0$
    \For{$i \gets 0$ to $(I_{l}-1)$}      
        \For{$j \gets 0$ to $(I_{b}-1)$}
            \State {$E^{G}[i,j] \gets uint8(E^{s}[k])$}
            \Comment{Converting to 'uint8'}
            \State $k \gets k+1$
        \EndFor
    \EndFor

    \end{algorithmic}
    \end{algorithm}

\subsubsection{Reshaping and Quantization}
\label{picr}
A typical image has two dimensions as opposed to the one-dimensional normalized and scaled sequence that was created in the previous phase. Reshaping process is used to change this one-dimensional sequence into a two-dimensional shape as portrayed in Algorithm \ref{alg:rsem}. The dimensions of the image matrix are predefined subject to the following constraint:
    \newline
    \begin{flushleft}
    $I_{l} \times I_{b}=dim_{e}$ \hspace*{0pt}\hfill (18) 
    \end{flushleft}

where, $I_l$ denotes the image length, $I_b$ the image breadth and $dim_e$ the embedding dimension as-well-as the length of the scaled sequence. This guarantees that all images have the same dimensions. Finally, this matrix is utilized to generate the grayscale image. Here, the pixels are converted into an 8-bit unsigned int (or 'uint8') notation. Compared to the standard 32-bit floating point ('float32') notation, this aids in reducing the memory footprint. The generated image representations can now be applied to downstream NLP tasks.

\begin{figure}[h]
\centering
\includegraphics[width=0.6\linewidth, height=15cm]{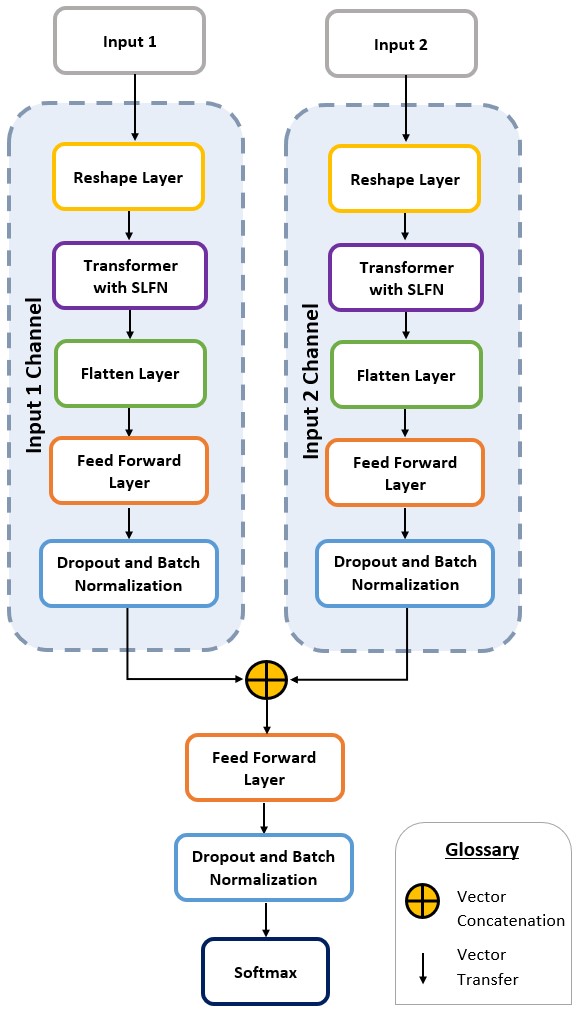}
\caption{Illustration of the proposed TSLFN-based STS model}
\label{fig11}       
\end{figure}

\subsection{Architecture for STS Determination}
\label{stsa}

To evaluate the efficacy of the TexIm FAST representations, a downstream STS task has been designed. For this, a novel transformer-based architecture has been devised for cross-modal determination of STS as illustrated in Figure \ref{fig11}. The architecture accepts two TexIm FAST images for comparison and outputs the similarity among them analogous to prevalent deep learning-based STS models in NLP \cite{Ref58, Ref61}. The inputs are propagated through two separate channels to obtain the hidden representations. Each channel includes a TSLFN block having architecture as shown in Figure \ref{fig12}. The hidden representations are then concatenated and processed further in the subsequent layers to predict their similarity. The proposed STS architecture is potent enough to analyze lengthy sequences despite being light-weight with less parameters. 

\section{Experiment Details}
\label{exp}
In this section, the experiment details such as the data-sets involved, implementation details as-well-as hyperparameter details of the proposed methodology have been enunciated.


\subsection{Data-Sets}
\label{data}

\begin{itemize}
    \item \textbf{Microsoft Research Paraphrase Corpus (MSRPC)}:
    It is a STS data-set released by Dolan and Brockett \cite{Ref65}. It comprises 5,801 pairs of sentences derived from news articles covering a range of topics. Each pair of sentences has been annotated as either similar or dissimilar in the form of a binary label. Out of the total pair of sentences, 3,900 pairs are paraphrases with the rest being dissimilar. The average length of the sentences is 18.92 words.
    
    \item \textbf{CNN/ Daily Mail}:
    It is an ATS data-set put forth by Nallapati et al. \cite{Ref63}. It comprises news articles with their summaries (highlights) retrieved from the CNN\footnote{https://www.cnn.com} and the Daily Mail websites\footnote{https://www.dailymail.co.uk}. For implementation on STS task, for 143,408 samples out of 286,817 sequences, the summaries have been shuffled. While for the rest, the summaries correspond to the original news. Additionally, a column has been added to denote whether the text and the summary are similar or not. For the news sequences, the overall length is 766 words distributed among 29.74 sentences. Whereas the highlights approximately contain 53 words over a span of 3.72 sentences.
    
    \item \textbf{XSum}:
    It has been curated by Narayan et al. \cite{Ref64} from BBC News\footnote{https://www.bbc.com/} for ATS. The data-set comprises articles along with the headlines as its summary. This data-set has been modified for STS such that 102,022 sequences have summaries corresponding to the given news and 102,023 sequences have the summaries shuffled. It consists of an additional column to indicate the similarity between the news article and its summary. Overall, the articles are 431.07 words spanning 19.77 sentences. Besides, the headlines consist of 23.26 words and are single sentences.
    
\end{itemize}

\subsection{Implementation Details}
\label{idet}
The methodologies proposed in this paper have been implemented on a Python 3 Compute Engine with Nvidia T4 GPU, 32 GB RAM, and 108 GB Disk in Google Colab\footnote{https://colab.research.google.com}. The maximum sequence lengths of the inputs corresponding to the MSRPC, CNN/ Daily Mail, and XSum data-sets are 64, 768, and 512 respectively. The output representations are grayscale images of size (32$\times$16)=512 pixels.

\subsection{Hyperparameter Details}
\label{hdet}
The hyperparameters of both the TexIm FAST model and the STS model have been finalized through Random Search based hyperparameter optimization. The final set of hyperparameters for the TexIm FAST and the STS model has been presented in Table \ref{table:table5} and Table \ref{table:table6} respectively.

\begin{table}[h!]
  \begin{center}
    \caption{Hyperparameter Details of TexIm FAST}
    \label{table:table5}
    \setlength{\tabcolsep}{1em} 
    \renewcommand{\arraystretch}{1.5} 
 \begin{tabular}{|l|l|} 
 \hline
 Hyperparameter & Value\\ 
 \hline
 \#Transformer Blocks & 2 \\
 \# Attention Heads & 4 \\
 \#Convolutional Blocks & 2 \\
 Activation Function & GELU \\
 Dropout Ratio & 0.3 \\
 Optimizer & Adam \\
 Learning Rate & 0.01 \\
 Train-Validation Split Ratio & 80:20 \\
 Early Stopping Patience Value & 2 \\
 Batch Size & 16 \\
 Training Epochs & 15 \\
 \hline
\multicolumn{2}{l}{Note- GELU: Gaussian Error Linear Unit} \\
 
\end{tabular}
\end{center}
\end{table}

\begin{table}[h!]
  \begin{center}
    \caption{Hyperparameter Details of the STS Model}
    \label{table:table6}
    \setlength{\tabcolsep}{1em} 
    \renewcommand{\arraystretch}{1.5} 
 \begin{tabular}{|l|l|} 
 \hline
 Hyperparameter & Value\\ 
 \hline
 \# Transformer Blocks & 1 \\
 \# Attention Heads & 4 \\
 Activation Function & LeakyReLU \\
 Dropout Ratio & 0.3 \\
 Optimizer & Adam \\
 Loss Function & Binary Cross-Entropy \\
 Learning Rate & 0.01 \\
 Train-Validation-Test Split Ratio & 70:15:15 \\
 Early Stopping Patience Value & 5 \\
 Batch Size & 16 \\
 Training Epochs & 50 \\
 \hline
 \multicolumn{2}{l}{Note- Leaky RELU: Leaky Rectified Linear Unit} \\
 
\end{tabular}
\end{center}
\end{table}

\begin{table*}[h!]
  \begin{center}
    \caption{Demonstration of TexIm FAST}
    \label{table:table2}
    \setlength{\tabcolsep}{1em} 
    \renewcommand{\arraystretch}{1.2} 
 \begin{tabular}{|p{16.3cm}|}

 \hline
 
 \begin{wrapfigure}{r}{0.18\textwidth} 
    \centering
    \includegraphics[width=0.18\textwidth, height=4.2cm]{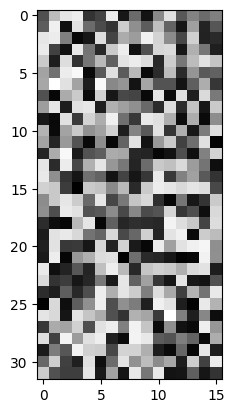}
    \caption{TexIm FAST representation}
    \label{fig3}
\end{wrapfigure}

 \textbf{Text:} 
 
 \textit{Ever noticed how plane seats appear to be getting smaller and smaller? With increasing numbers of people taking to the skies, some experts are questioning if having such packed out planes is putting passengers at risk. They say that the shrinking space on aeroplanes is not only uncomfortable- it's putting our health and safety in danger. More than squabbling over the arm rest, shrinking space on planes putting our health and safety in danger? This week, a U.S consumer advisory group set up by the Department of Transportation said at a public hearing that while the government is happy to set standards for animals flying on planes, it doesn't stipulate a minimum amount of space for humans. 'In a world where animals have more rights to space and food than humans,' said Charlie Leocha, consumer representative on the committee. 'It is time that the DOT and FAA take a stand for humane treatment of passengers.' But could crowding on planes lead to more serious issues than fighting for space in the overhead lockers, crashing elbows and seat back kicking? Tests conducted by the FAA use planes with a 31 inch pitch, a standard which on some airlines has decreased . Many economy seats on United Airlines have 30 inches of room, while some airlines offer as little as 28 inches. Cynthia Corbertt, a human factors researcher with the Federal Aviation Administration, that it conducts tests on how quickly passengers can leave a plane. But these tests are conducted using planes with 31 inches between each row of seats, a standard which on some airlines has decreased, reported the Detroit News. The distance between two seats from one point on a seat to the same point on the seat behind it is known as the pitch. While most airlines stick to a pitch of 31 inches or above, some fall below this. While United Airlines has 30 inches of space, Gulf Air economy seats have between 29 and 32 inches, Air Asia offers 29 inches and Spirit Airlines offers just 28 inches. British Airways has a seat pitch of 31 inches, while easyJet has 29 inches, Thomson's short haul seat pitch is 28 inches, and Virgin Atlantic's is 30-31. Experts question if packed out planes are putting passengers at risk. U.S consumer advisory group says minimum space must be stipulated.}
\\
 
 \hline
\end{tabular}
\end{center}
\end{table*}

\begin{figure*}[h!]
    \centering
    \begin{subfigure}[h]{0.15\textwidth}
        \centering
        \includegraphics[width=\linewidth, height=3cm]{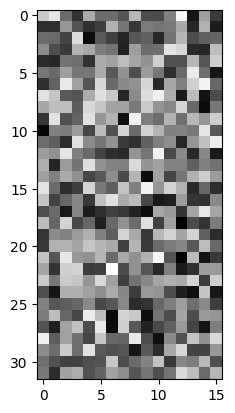}
        \caption{Sim\_MSRPC S1}
        \label{fig5a}       
    \end{subfigure}%
    ~ 
    \begin{subfigure}[h]{0.153\textwidth}
        \centering
        \includegraphics[width=\linewidth, height=3cm]{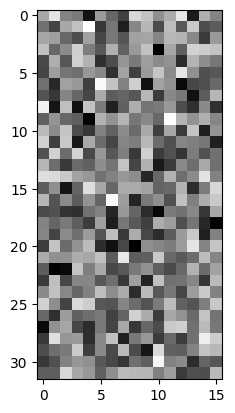}
        \caption{Dsim\_MSRPC S1}
        \label{fig5b}       
    \end{subfigure}
    ~
    \begin{subfigure}[h]{0.16\textwidth}
        \centering
        \includegraphics[width=\linewidth, height=3cm]{cnn_sim_text1.jpg}
        \caption{Sim\_CNNDM S1}
        \label{fig5c}       
    \end{subfigure}%
    ~
    \begin{subfigure}[h]{0.16\textwidth}
        \centering
        \includegraphics[width=\linewidth, height=3cm]{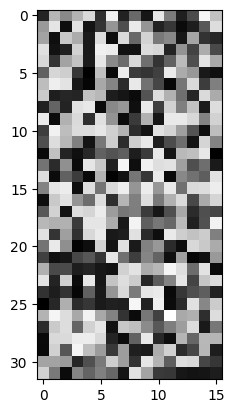}
        \caption{Dsim\_CNNDM S1}
        \label{fig5d}       
    \end{subfigure}%
    ~ 
    \begin{subfigure}[h]{0.15\textwidth}
        \centering
        \includegraphics[width=\linewidth, height=3cm]{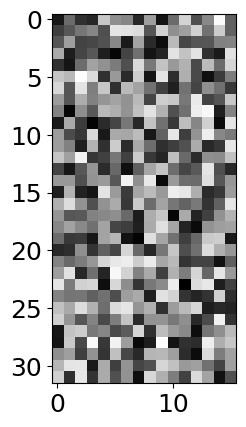}
        \caption{Sim\_XSum S1}
        \label{fig5e}       
    \end{subfigure}
    ~ 
    \begin{subfigure}[h]{0.15\textwidth}
        \centering
        \includegraphics[width=\linewidth, height=3cm]{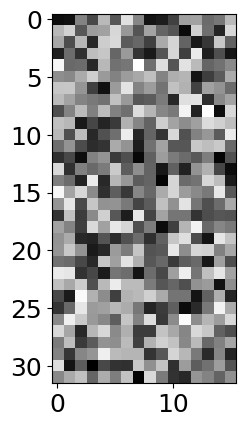}
        \caption{Dsim\_XSum S1}
        \label{fig5f}       
    \end{subfigure}
   
    \caption{TexIm FAST representations corresponding to the first sequences for all the comparison objectives}
    \label{fig5}
    
\end{figure*}

\begin{figure*}[h!]
    \centering
    \begin{subfigure}[h]{0.15\textwidth}
        \centering
        \includegraphics[width=\linewidth, height=3cm]{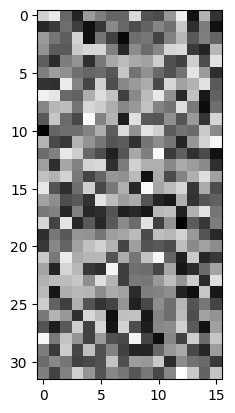}
        \caption{Sim\_MSRPC S2}
        \label{fig6a}       
    \end{subfigure}%
    ~ 
    \begin{subfigure}[h]{0.153\textwidth}
        \centering
        \includegraphics[width=\linewidth, height=3cm]{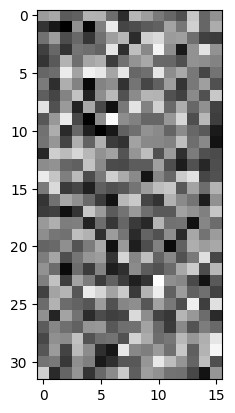}
        \caption{Dsim\_MSRPC S2}
        \label{fig6b}       
    \end{subfigure}
    ~
    \begin{subfigure}[h]{0.16\textwidth}
        \centering
        \includegraphics[width=\linewidth, height=3cm]{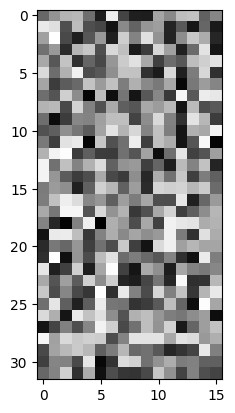}
        \caption{Sim\_CNNDM S2}
        \label{fig6c}       
    \end{subfigure}%
    ~
    \begin{subfigure}[h]{0.16\textwidth}
        \centering
        \includegraphics[width=\linewidth, height=3cm]{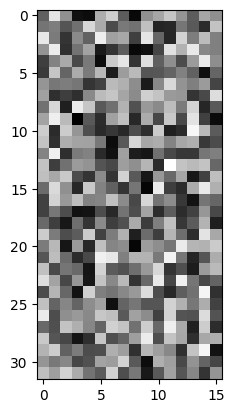}
        \caption{Dsim\_CNNDM S2}
        \label{fig6d}       
    \end{subfigure}%
    ~ 
    \begin{subfigure}[h]{0.15\textwidth}
        \centering
        \includegraphics[width=\linewidth, height=3cm]{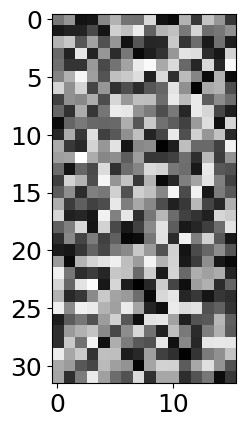}
        \caption{Sim\_XSum S2}
        \label{fig6e}       
    \end{subfigure}
    ~ 
    \begin{subfigure}[h]{0.15\textwidth}
        \centering
        \includegraphics[width=\linewidth, height=3cm]{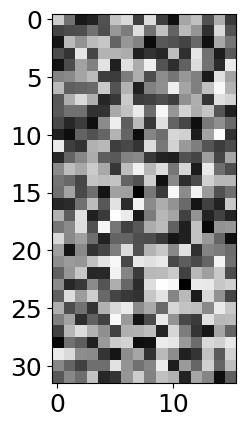}
        \caption{Dsim\_XSum S2}
        \label{fig6f}       
    \end{subfigure}
   
    \caption{TexIm FAST representations corresponding to the second sequences for all the comparison objectives}
    \label{fig6}
    
\end{figure*}

\section{Results and Discussion}
\label{rdis}
In this section, an extensive analysis of the results obtained from implementation of the proposed TexIm FAST has been performed on a variety of parameters and metrics.

\subsection{Demonstration of TexIm FAST}
\label{demo}
The text and the associated graphic representation created by TexIm FAST are shown in Table \ref{table:table2} and Figure \ref{fig3}, respectively, as part of the suggested methodology's demonstration. Each pixel in the visual representation represents a component of the low-dimensional sentence vector and expresses its value as pixel intensity. The pixels might not appear revealing enough at first glance. However, it incorporates the contextualized linguistic information from the text such that if two pictorial representations resemble one another, their provided texts must be semantically similar. 

\begin{figure*}[h!]
    \centering
    \begin{subfigure}[h]{0.30\textwidth}
        \centering
        \includegraphics[width=\linewidth, height=3cm]{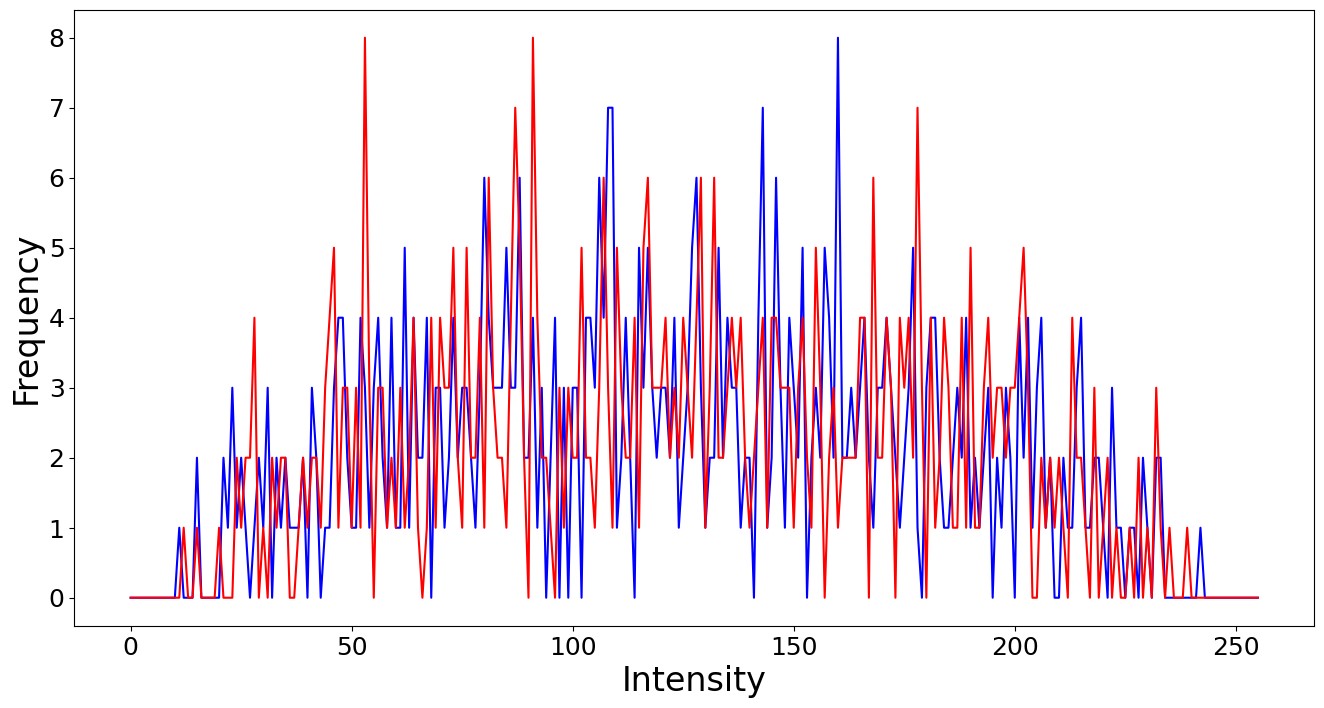}
        \caption{Sim\_MSRPC}
        \label{fig7a}       
    \end{subfigure}%
    ~ 
    \begin{subfigure}[h]{0.30\textwidth}
        \centering
        \includegraphics[width=\linewidth, height=3cm]{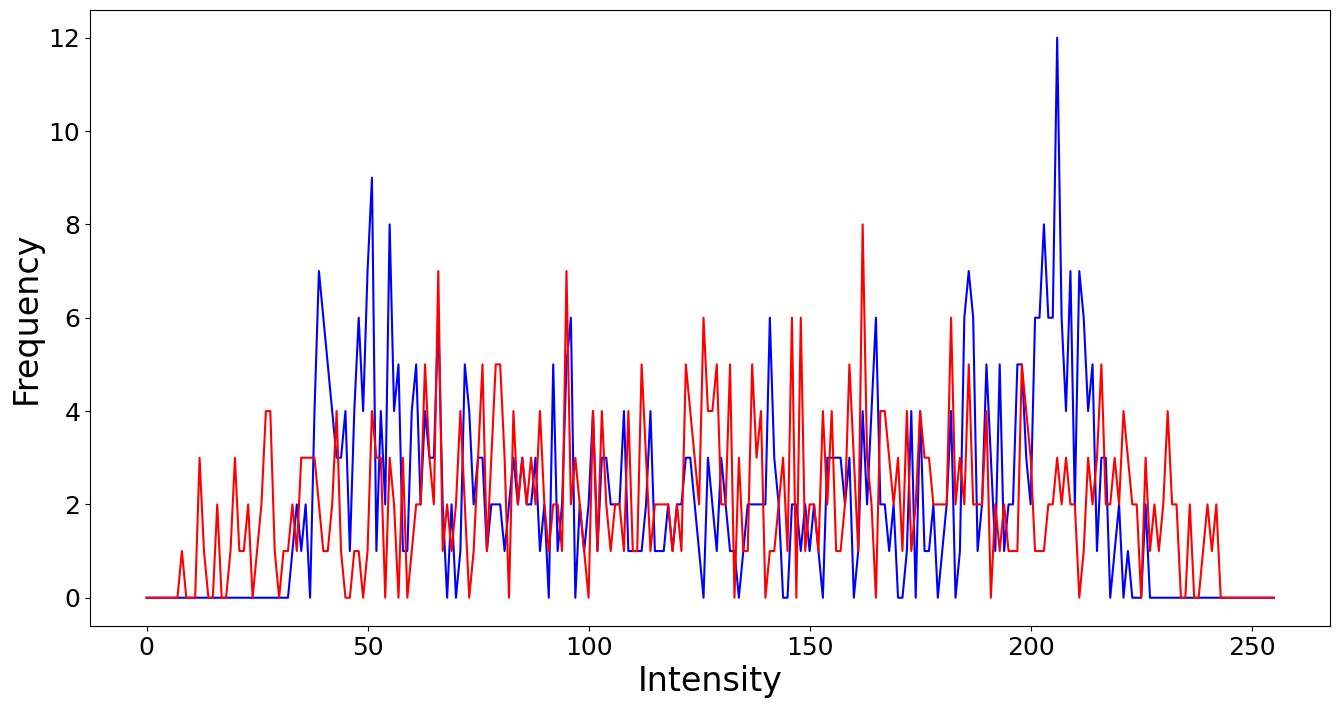}
        \caption{Sim\_CNNDM}
        \label{fig7c}       
    \end{subfigure}%
    ~
    \begin{subfigure}[h]{0.30\textwidth}
        \centering
        \includegraphics[width=\linewidth, height=3cm]{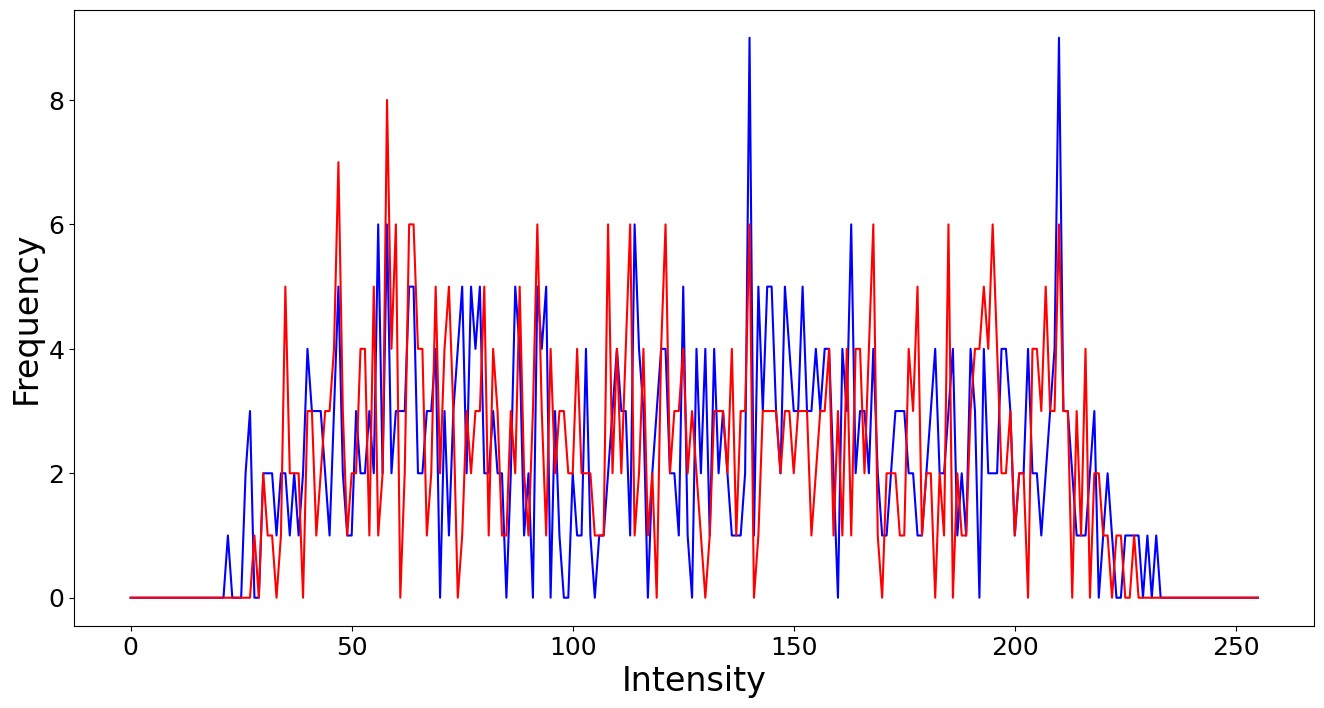}
        \caption{Sim\_XSum}
        \label{fig7e}       
    \end{subfigure}

    \bigskip
    
    \begin{subfigure}[h]{0.30\textwidth}
        \centering
        \includegraphics[width=\linewidth, height=3cm]{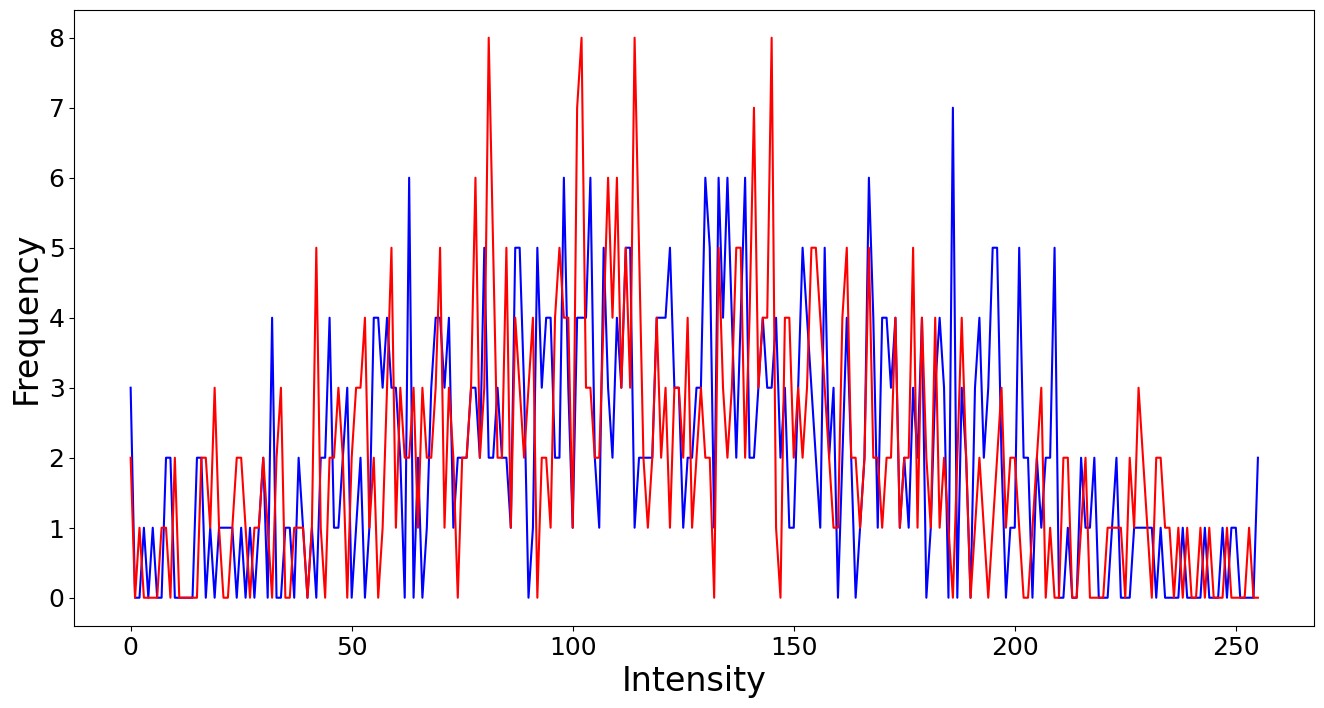}
        \caption{Dsim\_MSRPC}
        \label{fig7b}       
    \end{subfigure}
    ~
    \begin{subfigure}[h]{0.30\textwidth}
        \centering
        \includegraphics[width=\linewidth, height=3cm]{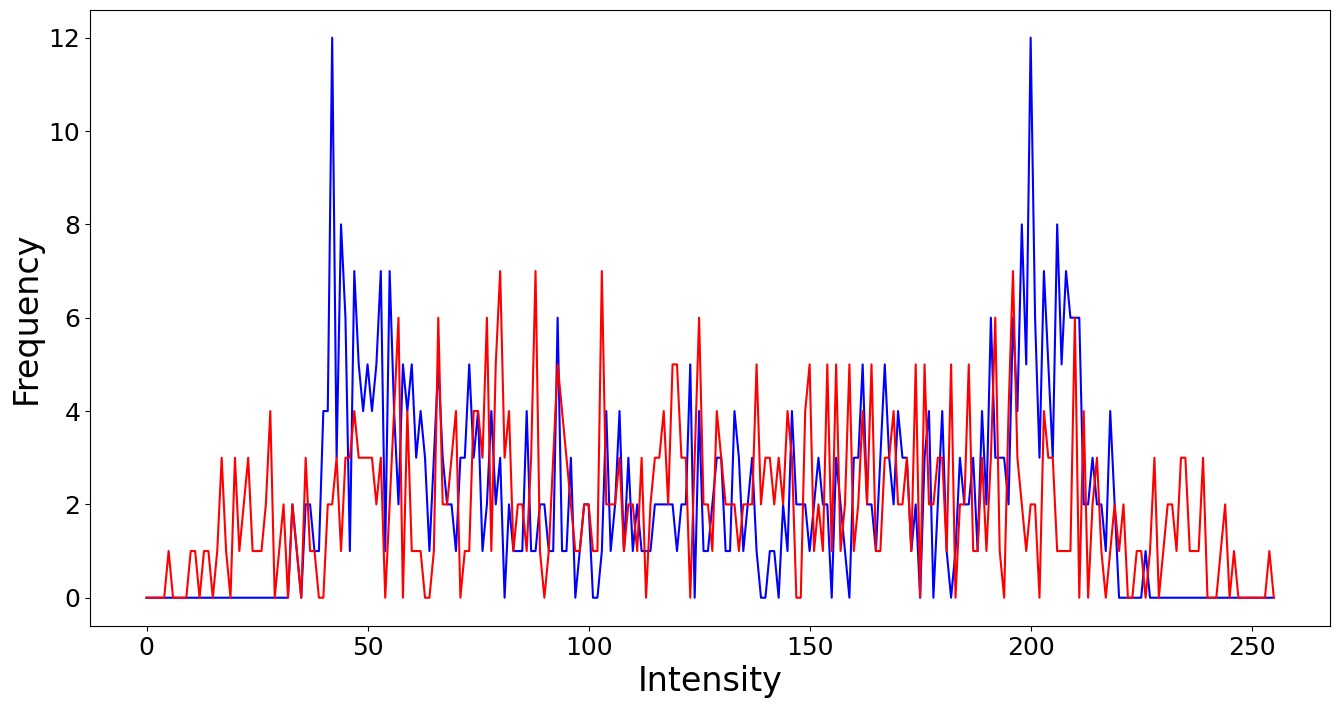}
        \caption{Dsim\_CNNDM}
        \label{fig7d}       
    \end{subfigure}%
    ~ 
    \begin{subfigure}[h]{0.30\textwidth}
        \centering
        \includegraphics[width=\linewidth, height=3cm]{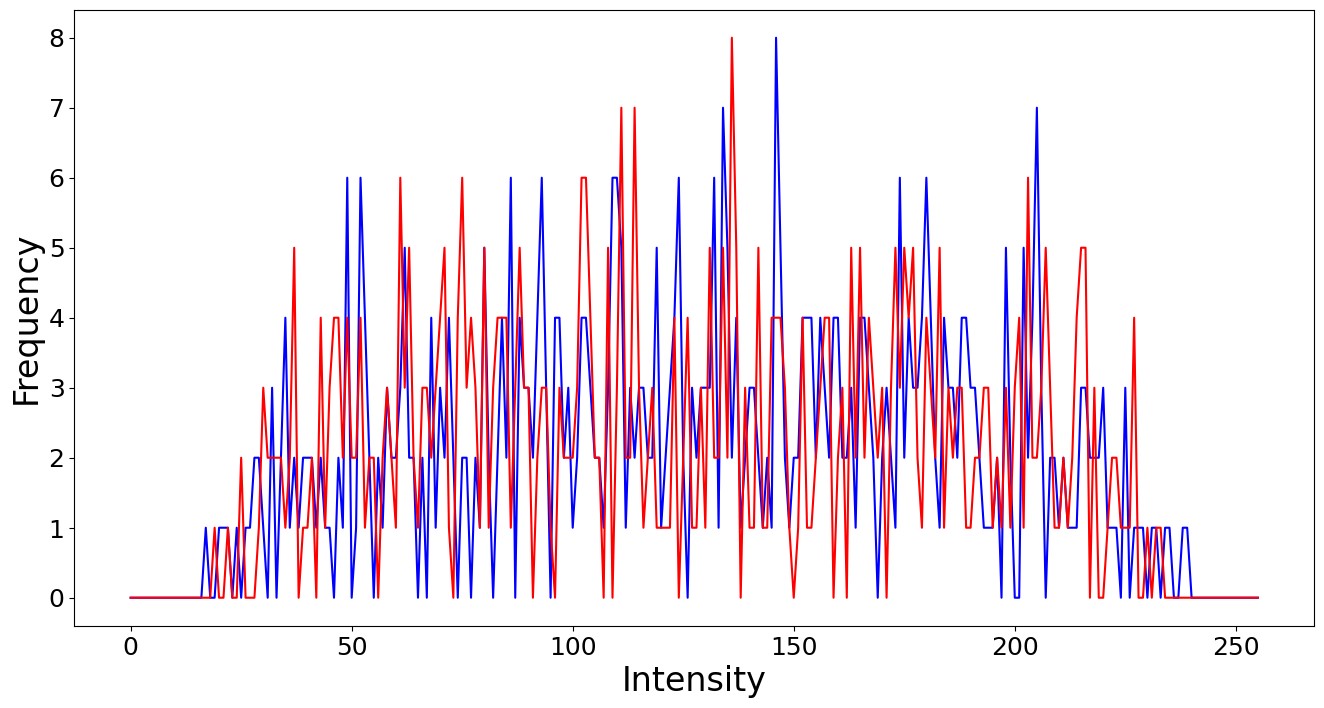}
        \caption{Dsim\_XSum}
        \label{fig7f}       
    \end{subfigure}
   
    \caption{Comparison of the histograms of the TexIm FAST representations for all the comparison objectives}
    \label{fig7}
    
\end{figure*}

\begin{figure*}[h!]
    \centering
    \begin{subfigure}[h]{0.30\textwidth}
        \centering
        \includegraphics[width=\linewidth, height=3cm]{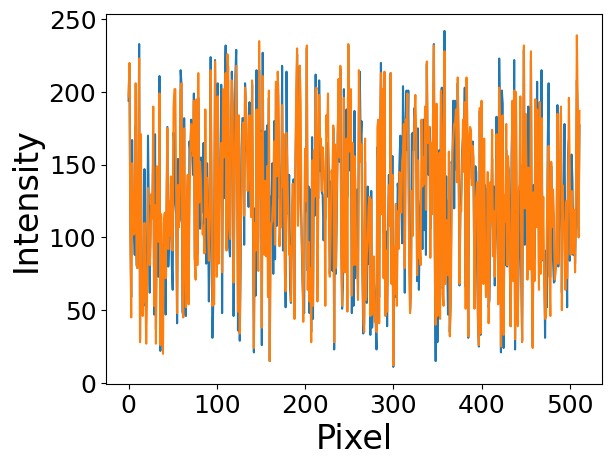}
        \caption{Sim\_MSRPC}
        \label{fig8a}       
    \end{subfigure}%
    ~ 
    \begin{subfigure}[h]{0.30\textwidth}
        \centering
        \includegraphics[width=\linewidth, height=3cm]{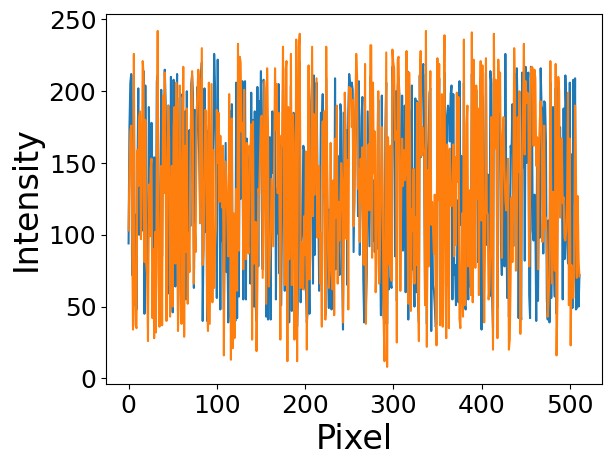}
        \caption{Sim\_CNNDM}
        \label{fig8c}       
    \end{subfigure}%
    ~ 
    \begin{subfigure}[h]{0.30\textwidth}
        \centering
         \includegraphics[width=\linewidth, height=3cm]{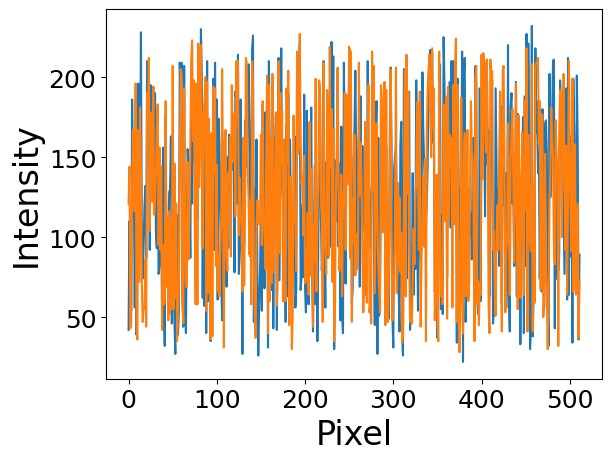}
        \caption{Sim\_XSum}
        \label{fig8e}       
    \end{subfigure}

    \bigskip
    
    \begin{subfigure}[h]{0.30\textwidth}
        \centering
        \includegraphics[width=\linewidth, height=3cm]{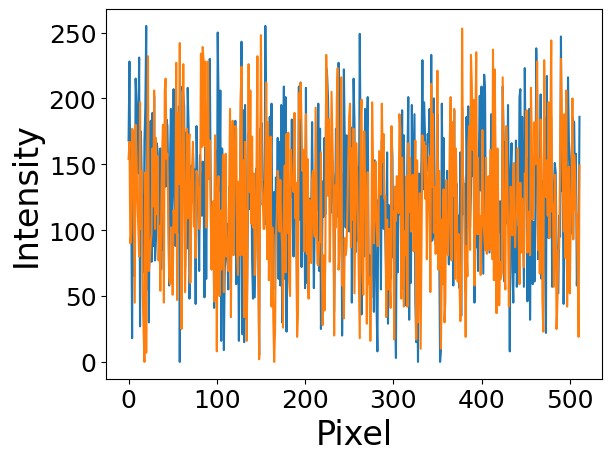}
        \caption{Dsim\_MSRPC}
        \label{fig8b}       
    \end{subfigure}
    ~
    \begin{subfigure}[h]{0.30\textwidth}
        \centering
        \includegraphics[width=\linewidth, height=3cm]{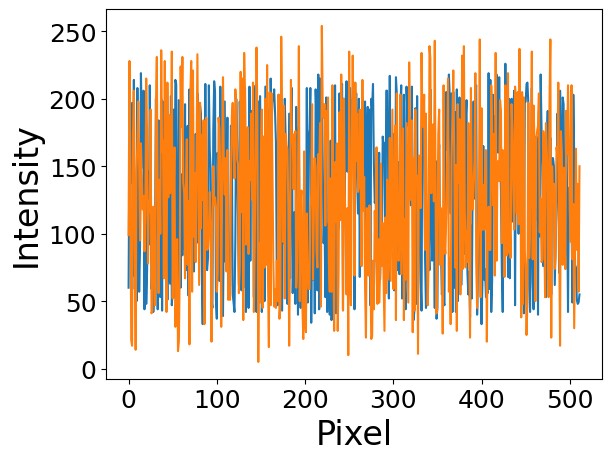}
        \caption{Dsim\_CNNDM}
        \label{fig8d}       
    \end{subfigure}%
    ~ 
    \begin{subfigure}[h]{0.30\textwidth}
        \centering
        \includegraphics[width=\linewidth, height=3cm]{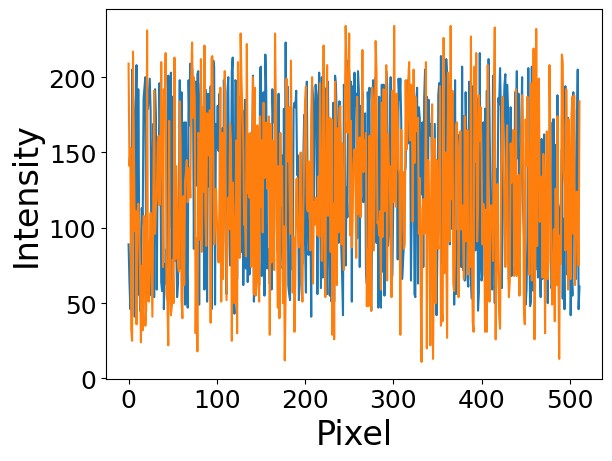}
        \caption{Dsim\_XSum}
        \label{fi8f}       
    \end{subfigure}
   
    \caption{Comparison of the line plots of the TexIm FAST representations for all the comparison objectives}
    \label{fig8}

\end{figure*}

\subsection{Comparison of Pictorial Representations}
To validate the capability of the pictorial representations generated by TexIm FAST in retaining the information in text, the following comparison objectives have been formulated:

\begin{enumerate}[(i)]
    \item \textbf{Comparing Similiar Texts:} The TexIm FAST representations corresponding to the pairs of semantically equivalent texts from MSRPC, CNN DailyMail, and XSum data sets have been compared giving rise to objectives \textit{Sim\_MSRPC}, \textit{Sim\_CNNDM} and \textit{Sim\_XSum} respectively. In \textit{Sim\_MSRPC}, the compared texts have similar lengths. Whereas in \textit{Sim\_CNNDM} and \textit{Sim\_XSum}, disparate length sequences involving a text and its summary have been analyzed.

    \item \textbf{Comparing Dissimilar Texts:} The TexIm FAST representations of to two semantically dissimilar texts have been compared to form the objectives \textit{Dsim\_MSRPC}, \textit{Dsim\_CNNDM} and \textit{Dsim\_XSum} corresponding to MSRPC, CNN DailyMail, and XSum data sets respectively. In \textit{Dsim\_MSRPC} two semantically dissimilar texts having similar lengths have been considered. While in \textit{Dsim\_CNNDM} and \textit{Dsim\_XSum}, a text and the summary of some other dissimilar text with considerably smaller length have been examined.

\end{enumerate}

Figure \ref{fig5} and Figure \ref{fig6} present the TexIm FAST representations of sequences for all the comparison objectives. Whereas, Figure \ref{fig7} illustrates the comparison of the histograms. Since a histogram only reflects the frequencies of individual intensities, Figure \ref{fig8} plots the pairwise pixel values for a better comparison between individual pixels. The results indicate that TexIm FAST captures the contextualized semantic information in the text and is potent enough to distinguish between similar and dissimilar texts. Furthermore, it is robust against the bias due to disparate sequence lengths as it effectively comprehends the (dis)similarity between long texts and concise summaries. This can be attributed to the ability to construct fixed-length representations from varying-length sequences.

\begin{table}[h!]
  \begin{center}
    \caption{Comparison of Performance}
    \label{table:table3}
    \setlength{\tabcolsep}{1em} 
    \renewcommand{\arraystretch}{1.5} 
 \begin{tabular}{|p{3cm}|p{3cm}|p{1.5cm}|p{2.1cm}|}
 \hline
 Data-Set & Method & Accuracy & F1-Score \\
 \hline
 MSRPC & BLEU \cite{Ref45} & 0.58 & 0.60 \\
    & ROUGE \cite{Ref46} & 0.69 & 0.67 \\
    & WMD \cite{Ref56} & 0.57 & 0.58 \\
    & S2SA-SNN \cite{Ref61} & 0.67 & 0.62 \\
    & TexIm V1 \cite{Ref11} & 0.68 & 0.66 \\
    & TexIm FAST-C  & 0.70 & 0.61 \\
   & \textbf{TexIm FAST} & \textbf{0.71} & \textbf{0.63} \\
 \hline
 CNN/ DailyMail & BLEU \cite{Ref45}  & 0.50 & 0.33\\
    & ROUGE \cite{Ref46} & 0.50 & 0.34 \\
    & WMD \cite{Ref56} & 0.70 & 0.70 \\
    & S2SA-SNN \cite{Ref61} & 0.52 & 0.52 \\
    & TexIm V1 \cite{Ref11} & 0.63 & 0.60 \\
    & TexIm FAST-C & 0.73 & 0.70 \\
   & \textbf{TexIm FAST} & \textbf{0.81} & \textbf{0.80} \\
\hline
 XSum & BLEU \cite{Ref45} & 0.50 & 0.33\\
    & ROUGE \cite{Ref46} & 0.52 & 0.35 \\
    & WMD \cite{Ref56} & 0.69 & 0.68 \\
    & S2SA-SNN \cite{Ref61} & 0.50 & 0.50 \\
    & TexIm V1 \cite{Ref11} & 0.60 & 0.57 \\
    & TexIm FAST-C  & 0.72 & 0.70 \\
   & \textbf{TexIm FAST} & \textbf{0.80} & \textbf{0.79} \\
 \hline
 
 \end{tabular}
 \end{center}
 \end{table}

\subsection{Performance Evaluation}
\label{peval}
The performance of the proposed TexIm FAST representations has been evaluated for the task of STS through comparison with the following works:

\begin{enumerate}[(i)]
    \item \textbf{ROUGE:} Recall-Oriented Understudy for Gisting Evaluation (ROUGE) metric proposed by Chin-Yew Lin \cite{Ref46} which relies upon the overlap of sub-sequences or n-grams to determining the similarity among a pair of texts.
    
    \item \textbf{BLEU:} Bilingual Evaluation Understudy (BLEU) metric proposed by Papineni et al. \cite{Ref45} to compute the similarity among two pieces of text. Unlike ROUGE, it is a precision-oriented metric with a penalty term to account for disparate sequence lengths.

    \item \textbf{WMD:} Word Mover Distance (WMD) by Kusner et al. \cite{Ref56} is a metric to ascertain the minimum distance for transportation of word embeddings from one document to the other. 

    \item \textbf{S2SA-SNN:} A Siamese Neural Network based on BiLSTM with cross-self attention proposed by Li et al. \cite{Ref61} for determining the semantic similarity.

    \item \textbf{TexIm V1:} Text-to-Image (TexIm) encoding technique for generating informed as-well-as memory efficient visual representation of text utilizing BERT-based contextualized embeddings \cite{Ref11}. These encodings are then fed into the proposed STS model.

    \item \textbf{TexIm FAST-C:} A variation of the proposed TexIm FAST devoid of the TSLFN block in the VAE, i.e. comprising only the convolutional blocks.
    
\end{enumerate}

From the comparison with related works in Table \ref{table:table3}, it can be observed that the BLEU and ROUGE falter in the case of long sequences as they are suited for short sequences. A contrasting observation is noted for WMD, where the accuracy rises while comparing corpus with lengthy sequences. This can be attributed to establishing semantic relatedness of similar words in both documents instead of matching sub-sequences. S2SA-SNN being constituted from word-level representations, provides satisfactory accuracy for MSRPC having similar-length sequences for comparison. But for data-sets having unequal-length sequences like CNN/ Daily Mail and XSum, the accuracy drops considerably. In the case of TexIm V1, the performance improves due to informed contextualized representations. But, due to the dependency of the representations on the sequence length, the accuracy declines when comparing disparate length sequences. The performance improves further with TexIm FAST-C architecture similar to the proposed TexIm FAST, distinguished by the absence of the TSLFN block. When TexIm FAST representations are fed into the proposed STS model, overall 5.6\% superior performance compared to TexIm FAST-C is obtained. Moreover, on the CNN/ Daily Mail and the XSum data-sets the improvement in accuracy is significantly high. This highlights the inability of existing approaches to deal with disparate length sequences and simultaneously the proficiency of the TexIm FAST representations as-well-as the proposed STS model to overcome this drawback.

\subsection{Ablation Study}
\label{abls}
The effectiveness of the parameters used for the proposed TexIm FAST and the generated represenatations has been validated through a detailed ablation study.

\subsubsection{Image Dimensionality and Number of Channels}
\label{idim}
The dimensionality as-well-as the number of channels of the proposed TexIm FAST has been selected after analyzing the effectiveness in terms of information retained as-well-as efficiency in terms of memory footprint of the images. For this, various image dimensions and channel combinations have been studied as follows:

 \begin{figure}[h!]
    \centering
    \begin{subfigure}[h]{0.15\linewidth}
        \centering
        \includegraphics[width=\linewidth, height=2.4cm]{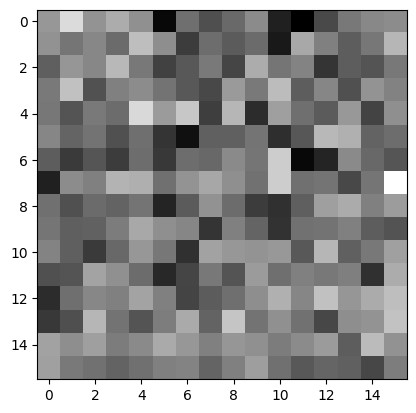}
        \caption{Grayscale~256}
        \label{fig9a}       
    \end{subfigure}%
    ~ 
    \begin{subfigure}[h]{0.24\linewidth}
        \centering
        \includegraphics[width=\linewidth, height=4.2cm]{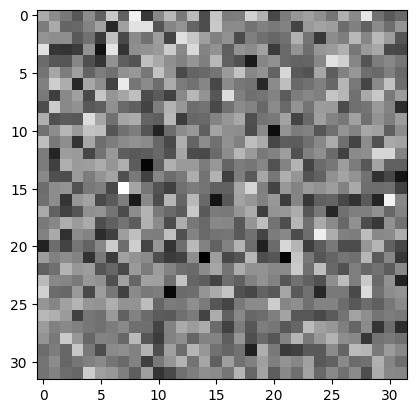}
        \caption{Grayscale 1024}
        \label{fig9b}       
    \end{subfigure}
    ~
    \begin{subfigure}[h]{0.18\linewidth}
        \centering
        \includegraphics[width=\linewidth, height=4.2cm]{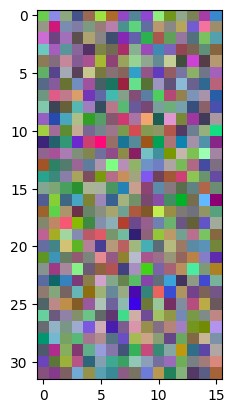}
        \caption{RGB 512}
        \label{fig9c}       
    \end{subfigure}%
    \caption{Sample representations generated by the model ablations upon the MSRPC Data-Set}
    \label{fig9}

\end{figure}

\begin{figure*}[h!]
    \centering
    \begin{subfigure}[h]{0.321\linewidth}
        \centering
        \includegraphics[width=\linewidth]{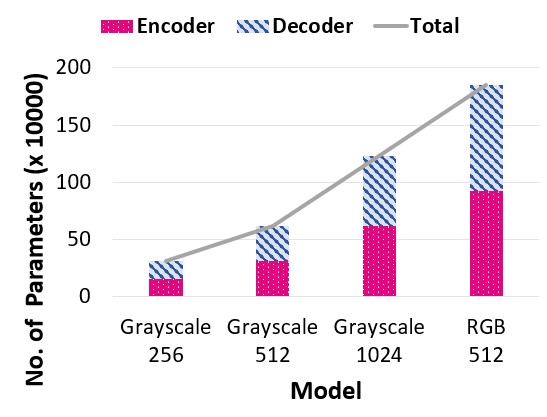}
        \caption{MSRPC Data-Set}
        \label{fig10a}       
    \end{subfigure}%
    ~
    \begin{subfigure}[h]{0.321\linewidth}
        \centering
        \includegraphics[width=\linewidth]{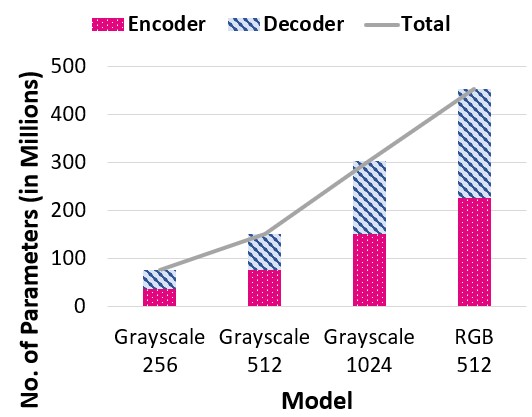}
        \caption{CNN/ Daily Mail Data-Set}
        \label{fig10b}       
    \end{subfigure}
    ~
    \begin{subfigure}[h]{0.321\linewidth}
        \centering
        \includegraphics[width=\linewidth]{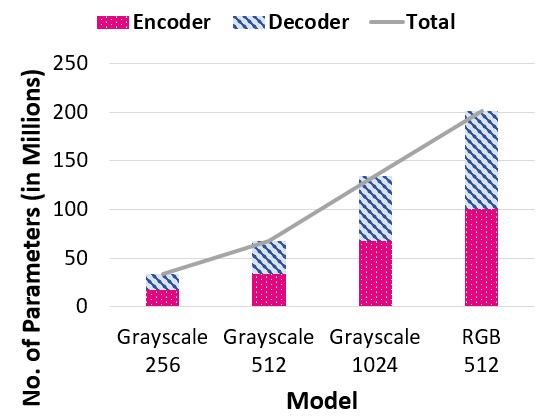}
        \caption{XSum Data-Set}
        \label{fig10c}       
    \end{subfigure}%
    \caption{Comparison of the number of model parameters for ablations of TexIm FAST models upon various data-sets}
    \label{fig10}

\end{figure*}
 
\begin{enumerate}[(i)]
    \item \textbf{Grayscale 256}: A grayscale image having 256 pixels having length as 16 pixels and width as 16 pixels (specimen Figure \ref{fig9a}).

    \item \textbf{Grayscale 512 (Proposed Methodology)}: A grayscale image having 32 $\times$ 16 pixels, i.e. 512 pixels. It is the selected dimensionality of the proposed TexIm FAST.

    \item \textbf{Grayscale 1024}: A grayscale image having 32 $\times$ 32 pixels, i.e. 1024 pixels (specimen Figure \ref{fig9b}).

    \item \textbf{RGB 512}: RGB representations with three channels having 32 $\times$ 16 $\times$ 3 pixels, i.e. 1536 pixels (specimen Figure \ref{fig9c}).
    
\end{enumerate}

\begin{table}[h!]
  \begin{center}
    \caption{Analysis of Variation in Image Dimensionality and Number of Channels}
    \label{table:table1}
    \setlength{\tabcolsep}{1em} 
    \renewcommand{\arraystretch}{1.5} 
 \begin{tabular}{|p{3cm}|p{2.7cm}|p{1.2cm}|p{1.5cm}|}
 \hline
 Data-Set & Model & Accuracy & F1-Score \\
 \hline
MSRPC & Grayscale 256 & 0.66 & 0.55 \\
 & Grayscale 512 & \textbf{0.71} & 0.63 \\
 & Grayscale 1024 & 0.68 & 0.62 \\
 & RGB 512 & 0.69 & \textbf{0.64} \\
  \hline
CNN/ Daily Mail & Grayscale 256 & 0.76 & 0.75 \\
 & Grayscale 512 & 0.81 & 0.80 \\
 & Grayscale 1024 & 0.82 & 0.82 \\
 & RGB 512 & \textbf{0.86} & \textbf{0.86} \\
  \hline
Xsum & Grayscale 256 & 0.77 & 0.75 \\
 & Grayscale 512 & 0.81 & 0.80 \\
 & Grayscale 1024 & 0.83 & 0.83 \\
 & RGB 512 & \textbf{0.85} & \textbf{0.85} \\

 \hline
 
 \end{tabular}
 \end{center}
 \end{table}

Table \ref{table:table1} depicts the comparison of accuracy across the model ablations varying in the image dimensionality and number of channels while Figure \ref{fig10} compares the number of parameters of the ablations. Here, it can be observed that with increasing dimensions the informativeness of the representations increases along with a decrease in memory efficiency. For a given image dimensionality, the shorter sequences are more informative compared to long sequences. Besides, the RGB representations provide the highest accuracy for the CNN/ Daily Mail and XSum data-sets comprising long sequences. For the MSRPC data-set with short sentences, while the F1-score is the highest, not much improvement can be noticed in terms of accuracy. This can be attributed to the fact that as the sequences are condensed into fixed-length representations, for shorter sequences more information can be packed while some information is lost in the case of long sequences. However, excessively larger images might not be effective beyond a certain extent due to information redundancy and rise in memory consumption. Furthermore, Figure \ref{fig10} depicts that the number of model parameters rises with increasing resolution and number of channels. Therefore, the rise in performance comes at the cost of efficiency.

\bigskip
\subsubsection{Contribution of TexIm in STS}
\label{ets}
After analyzing the efficacy of the variations of TexIm FAST representations, the contribution of the TexIm encodings as a whole towards STS has been studied. For this, the performance of the proposed STS model is examined through the following ablations:

\begin{table}[h!]
  \begin{center}
    \caption{Analysis of the Contribution of TexIm in STS}
    \label{table:table7}
    \setlength{\tabcolsep}{1em} 
    \renewcommand{\arraystretch}{1.5} 
 \begin{tabular}{|p{3cm}|p{2.7cm}|p{1.2cm}|p{1.5cm}|}
 \hline
 Data-Set & Method & Accuracy & F1-Score \\
 \hline
 MSRPC & Discrete STS & 0.67 & 0.65 \\
   & Tex FAST & 0.72 & 0.66 \\
   &  TexIm FAST &  0.71 &  0.63 \\
 \hline
 CNN/ DailyMail & Discrete STS & 0.59 & 0.57 \\
   & Tex FAST & 0.81 & 0.81 \\
   &  TexIm FAST &  0.81 &  0.80 \\
\hline
 XSum & Discrete STS & 0.57 & 0.54 \\
   & Tex FAST & 0.81 & 0.80 \\
   &  TexIm FAST &  0.80 &  0.79 \\
 \hline
 
 \end{tabular}
 \end{center}
 \end{table}

\begin{enumerate}[(i)]
    \item \textbf{Discrete STS:} This version comprises of the STS block isolated from the proposed TexIm FAST model. Here, the pre-processed input text is fed directly to the proposed STS model without any other transformations. 

    \item \textbf{Tex FAST}: This is an ablation of the TexIm FAST wherein instead of generating the visual representations, the fixed-length encodings obtained from the proposed TexIm FAST model are directly fed into the STS model.
    
\end{enumerate}

From the analysis of model ablations related to the inclusion of TexIm representations in Table \ref{table:table7}, it can be inferred that Discrete STS falters in performance compared to TexIm FAST as the model input is devoid of the linguistic characteristics and contextual information imbibed in the TexIm FAST encodings. Additionally, the performance drops significantly for disparate-length sequences in the case of CNN/ Daily Mail and XSum data-sets due to absence of the fixed-length encodings like TexIm FAST. Besides, the Tex FAST slightly excels the proposed TexIm FAST. This difference can be attributed to the loss of information due to the quantization performed in TexIm FAST. However, on estimating the accuracy versus efficiency trade-off in \textit{Section \ref{eeval}}, this $\approx$ 1 \% drop in accuracy reduces the memory consumption by 75\%. 

\begin{table*}[h!]
  \begin{center}
    \caption{Comparison of Compression Capability}
    \label{table:table4}
    \setlength{\tabcolsep}{1em} 
    \renewcommand{\arraystretch}{1.5} 
 \begin{tabular}{|p{4.5cm}|p{3.9cm}|p{2cm}|p{3cm}|}
 \hline
 Representation & Conventional Approach & TexIm FAST & Compression Achieved \\
 \hline
 Plain text representation$^{1}$ & 2123.57B & 512B & 75.89\% \\
 Word Embedding (Word2Vec)$^{2}$ & 61440B & 512B & 99.92\% \\
 Word Embedding (BERT) $^{3}$ & 1572864B & 512B & 99.97\% \\
 Sequence Embedding $^{4}$ & 2048B & 512B & 75\% \\
 TexIm \cite{Ref11} & 1536B & 512B & 66.67\% \\
 \hline
 \multicolumn{4}{p{15cm}}{$^{1}$ Average memory required to store each sequence in the data-sets studied in this paper. \newline
$^{2}$ Memory required to store a 512 length sequence using 300 dimensional Word2Vec word embeddings. \newline
$^{3}$ Memory required to store a 512 length sequence using 768 dimensional BERT word embeddings. \newline
$^{4}$ Memory required to store a 512 length sequence using sequence embedding having 512 dimensions.} \\
 \end{tabular}
 \end{center}
 \end{table*}

\subsection{Efficiency Evaluation}
\label{eeval}
Comparing the efficiency in terms of model parameters as illustrated in Figure \ref{fig10}, it can be inferred that a rise in the image dimensionality and number of channels contribute to escalation in the model parameters leading to a decline in efficiency. For the MSRPC data-set, the number of parameters is the lowest while for the CNN/ Daily Mail the model parameters are the highest. Thus, the number of model parameters is proportional to the length of input sequences. 

In terms of memory compression, the efficiency of the TexIm FAST representations compared to conventional text representation techniques has been analyzed in Table \ref{table:table4}. Compared to plain text, the TexIm FAST representations provide compression for any sequences longer than 107 words. This is in consideration of the fact that in the English language, a word contains approximately 4.79 characters\footnote{http://www.norvig.com/mayzner.html}. For word embedding techniques, the compression achieved is over 99\%. For sequence embeddings, although the dimensionality of the vector remains the same, the compression is due to the application of unsigned int ('uint8') notation in TexIm FAST in place of 32-bit float ('float32') notation adopted in conventional embeddings. Moreover, over 66\% compression is achieved compared to TexIm \cite{Ref11} due to single-channel grayscale representations in place of three-channel RGB representations. Overall, it can be observed that the proposed TexIm FAST achieves above 75\% compression in terms of memory footprint compared to the conventional representation techniques. 

\subsection{Applications of TexIm FAST}
\label{app}
Apart from STS, the TexIm FAST representations are potent enough to be applied to a wide variety of downstream tasks. In spite of being indecipherable to the human eye, the representations can serve as input to models for further analysis enabling oblivious inference. This makes it appropriate for applications involving sensitive data such as patient records, customer repository and defence related data. In such applications, the TexIm FAST representations can be stored in the servers and used for further processing instead of performing the operations on the original data to ensure privacy. Moreover, the TexIm FAST representations are irreversible. Thus, even if some malicious agent obtains these representations, the actual text cannot be recovered from it. Furthermore, TexIm FAST significantly reduces the input memory requirements for downstream tasks making it deployable to devices with limited computational power.

\section{Conclusion}
\label{cocl}
This paper augments the body of knowledge through a self-supervised architecture titled TexIm FAST for generating fixed-length pictorial representations from text. The representations allow oblivious inference by retaining the linguistic intricacies of the original text along with a reduction in the memory footprint by over 75\%. Furthermore, a novel transformer encoder architecture has been formulated for STS. The efficacy of the proposed TexIm FAST has been analyzed through an extensive set of experiments. The results demonstrate that the proposed methdologies excel in the task of STS, and deliver noteworthy performance when dealing with disparate length sequences. However, encoding variable-length sequences to fixed-length representations implies that for smaller sequences the amount of information retained is higher and reduces as the sequence length increases. Therefore, an adaptive approach can be devised to determine the image resolution based on the sequence length. In the future, the potential of TexIm FAST can be examined on a wide range of NLP tasks as-well-as image processing models. This paper ushers a new avenue of text-to-image representation, with ample scope for formulating more effective as-well-as efficient techniques for cross-modal analysis of text.

\section*{Acknowledgments}
None. The author(s) received no financial or any other kinds of support for the research, authorship, and/or publication of this article.


\end{document}